\def\CLASSINPUTinnersidemargin{0.75in}
\def\CLASSINPUToutersidemargin{0.75in}
\def\CLASSINPUTtoptextmargin{0.75in}
\def\CLASSINPUTbottomtextmargin{0.75in}
\documentclass[conference]{IEEEtran}
\IEEEoverridecommandlockouts
\def\IEEEtitletopspaceextra{10pt}
\usepackage{amsmath,amsfonts}
\usepackage{algorithmic,algorithm}
\usepackage{array}
\usepackage[caption=false,font=normalsize,labelfont=sf,textfont=sf]{subfig}
\usepackage{textcomp}
\usepackage{stfloats}
\usepackage{url}
\usepackage{verbatim}
\usepackage{graphicx}
\usepackage{booktabs}
\usepackage{hyperref}
\hypersetup{colorlinks=true, linkcolor=blue, filecolor=black, urlcolor=blue, citecolor=blue}
\usepackage[nameinlink]{cleveref}
\crefname{figure}{Figure}{Figures}
\crefname{algorithm}{Algorithm}{Algorithms}
\crefname{table}{Table}{Tables}

\usepackage{placeins}
\graphicspath{{./Pictures/}}
\usepackage[switch]{lineno}

\def\BibTeX{{\rm B\kern-.05em{\sc i\kern-.025em b}\kern-.08em
    T\kern-.1667em\lower.7ex\hbox{E}\kern-.125emX}}

\usepackage{balance}

\begin{document}
\title{RoMu4o: A Robotic Manipulation Unit For Orchard Operations Automating Proximal Hyperspectral Leaf Sensing}





\author{
\IEEEauthorblockN{Mehrad Mortazavi$^{1}$, David J. Cappelleri$^{2}$, Reza Ehsani$^{1}$} 

\thanks{This work was supported by the IoT4Ag Engieering Research Center supported by the NSF under the cooperative agreement No. EEC-1941529.

$^{1}$ Mehrad Mortazavi and Reza Ehsani are with the Department of Mechanical Engineering, University of California, Merced, CA, 95343, USA. {Email: \{smortazavi3, rehsani\}@ucmerced.edu}

$^{2}$ David J. Cappelleri is with the School of Mechanical Engineering and Weldon School of Biomedical Engineering (By Courtesy), Purdue University, West Lafayette, IN USA 47907. Email: dcappell@purdue.edu

The code repository is accessible on \href{https://github.com/mehradmrt/UCM-AgBot-ROS2}{GitHub}.
}
}

\markboth{}%
{How to Use the IEEEtran \LaTeX \ Templates}

\maketitle

\begin{abstract}
Driven by the need to address labor shortages and meet the demands of a rapidly growing population, robotic automation has become a critical component in precision agriculture. Leaf-level hyperspectral spectroscopy is shown to be a powerful tool for phenotyping, monitoring crop health, identifying essential nutrients within plants as well as detecting diseases and water stress. This work introduces RoMu4o, a robotic manipulation unit for orchard operations offering an automated solution for proximal hyperspectral leaf sensing. This ground robot is equipped with a 6DOF robotic arm and vision system for real-time deep-learning based image processing and motion planning. We developed robust perception and manipulation pipelines that enable the robot to successfully grasp target leaves and perform spectroscopy. These frameworks operate synergistically to identify and extract the 3D structure of leaves from an observed batch of foliage, propose 6D poses, and generate collision-free constraint-aware paths for precise leaf manipulation. The end-effector of the arm features a compact design that integrates an independent lighting source with a hyperspectral sensor, enabling high-fidelity data acquisition while streamlining the calibration process for accurate measurements. Our ground robot is engineered to operate in unstructured orchard environments. However, the performance of the system is evaluated in both indoor and outdoor plant models. The system demonstrated reliable performance for 1-LPB hyperspectral sampling, achieving 95\% success rate in lab trials and 79\% in field trials. Field experiments revealed an overall success rate of 70\% for autonomous leaf grasping and hyperspectral measurement in a pistachio orchard.

\end{abstract}

\IEEEoverridecommandlockouts

\newcommand{\cprfont}{\footnotesize}                     
\newlength{\cprwidth}\setlength{\cprwidth}{\textwidth}   
\newlength{\cprshift}\setlength{\cprshift}{2.5\baselineskip}  
\setlength{\fboxrule}{0.5pt}                             
\setlength{\fboxsep}{4pt}                                

\IEEEpubid{%
  \raisebox{-\cprshift}[0pt][0pt]{%
    \fbox{\parbox{\cprwidth}{\cprfont
      \copyright~2026 IEEE. Personal use of this material is permitted.
      Permission from IEEE must be obtained for all other uses, in any current
      or future media, including reprinting/republishing this material for
      advertising or promotional purposes, creating new collective works, for
      resale or redistribution to servers or lists, or reuse of any copyrighted
      component of this work in other works.}}%
  }%
}


\section{Introduction}
The future of agriculture is being progressively shaped by the innovation of novel robotic platforms that transform key agricultural practices. Unmanned ground vehicles (UGVs) are an integral component of a smart farm and have been extensively used for research and commercial purposes in agriculture. Agricultural UGVs are developed and used for crop monitoring and physical sampling \cite{kim2022p}, harvesting \cite{zhou2022intelligent,yoshida2022fruit}, seeding and planting \cite{blender2016managing}, weed detection and removal \cite{albani2017monitoring}, precision irrigation and fertilization \cite{thayer2018multi}, phenotyping \cite{gao2018novel}, and pruning \cite{you2023semiautonomous}. Innovation of new agricultural robots is critical globally, particularly in countries facing severe labor shortages. Consistent performance under harsh conditions, cost reduction, and safety improvement in the long-term have been major reasons for the agricultural industry to deploy autonomous robotic systems in field operations \cite{wang2021smart}.

\begin{figure}[t]
    \centering
    \includegraphics[width=1\linewidth]{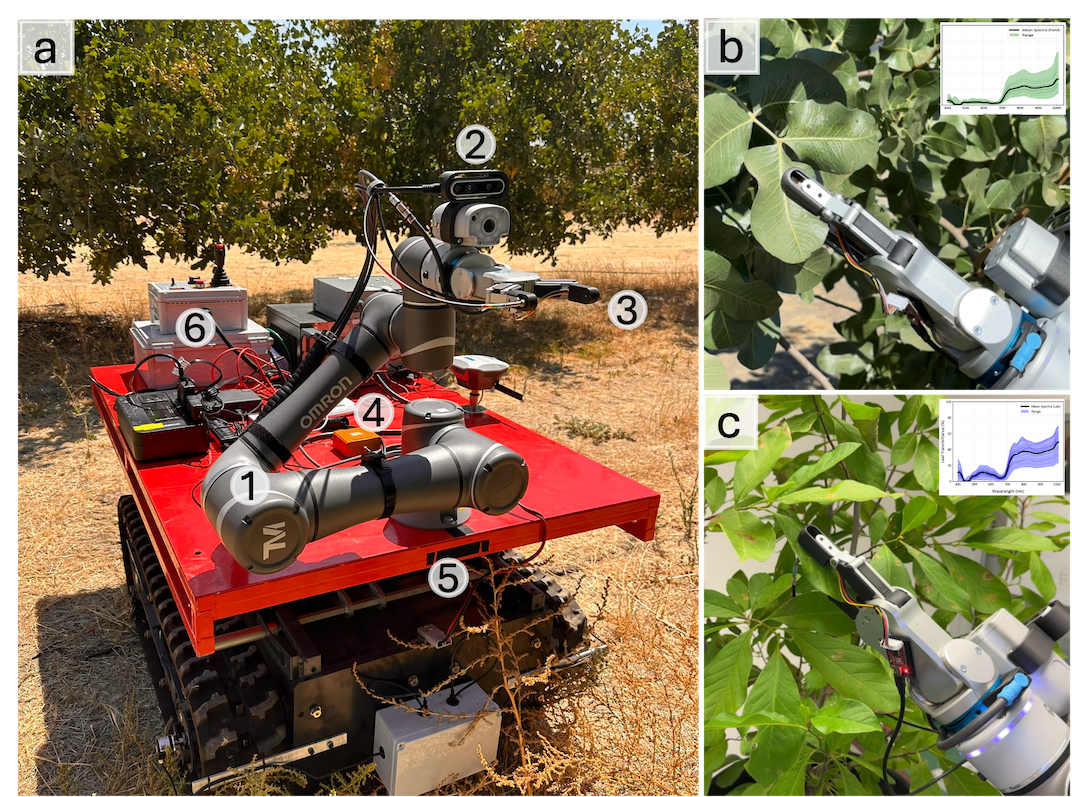}
    \caption{\textbf{(a)} RoMu4o, our proposed robotic manipulation unit for orchard operations aimed for autonomous proximal hyperspectral leaf sensing. The system is comprised of 1) 6-DOF robotic manipulator 2) RGBD camera 3) two-finger gripper end-effector integrated with a hyperspectral sensing system 4) an independent VIS-NIR light source for referencing and self-calibration 5) crawler-type ground robot 6) electric and control boxes, hardware drivers, portable batteries, processing unit, Ethernet hub, and auxiliary hardware needed for the UGV's navigation. \textbf{(b)} Pistachio leaves used for field experiments; \textbf{(c)} Magnolia leaves used for lab experiments.}
    \label{fig:system_design}
\end{figure}

Persistent crop monitoring is an essential task to mitigate crop loss and maximize yield output \cite{lei20244d}. It is crucial to identify plant biotic and abiotic stress at an early stage when severe damage to crops is preventable. Leaf-level hyperspectral imaging is proven to be effective in the detection of nutrient deficiencies, diseases, and water stress \cite{xu2022review,junttila2022close}. Hyperspectral data can be obtained remotely via drones and satellites or at the leaf level with handheld devices. Remote sensing can provide data at scale but requires rigorous calibration due to susceptibility to the outdoor lightning conditions. In contrast, portable hyperspectral spectroradiometers benefit from an independent lighting source that streamlines the calibration process and effectively reduces the sensitivity of the image to outdoor environmental conditions. However, there is a trade-off between high-fidelity data, and operational scalability, as manual hyperspectral data collection is time- and labor-intensive. This challenge underscores the potential for developing a robotic manipulation system equipped with a portable hyperspectral sensor capable of delivering high-quality data while reducing labor demands and bridging the gap between data fidelity and scalability. Automating leaf-level hyperspectral imaging with a robotic manipulator faces specific challenges. Unstructured canopies introduce uncertainty in leaf detection, and crop manipulation \cite{Kim2024deformable}. Dynamic lighting conditions significantly affect the performance of the sensors and computer vision algorithms \cite{lee2024towards, kim2025autonomous}. Elevated ambient temperatures can adversely affect the performance of an integrated hyperspectral sensor. Variable foliage density and hidden branches raise the risk of collision. Furthermore, in real environments, leaf displacement caused by small noise such as a light breeze can compromise the success rate of robotic manipulation.

\begin{figure*}[t]
    \centering
    \includegraphics[width=1\linewidth, trim={0 50 0 0}, clip]{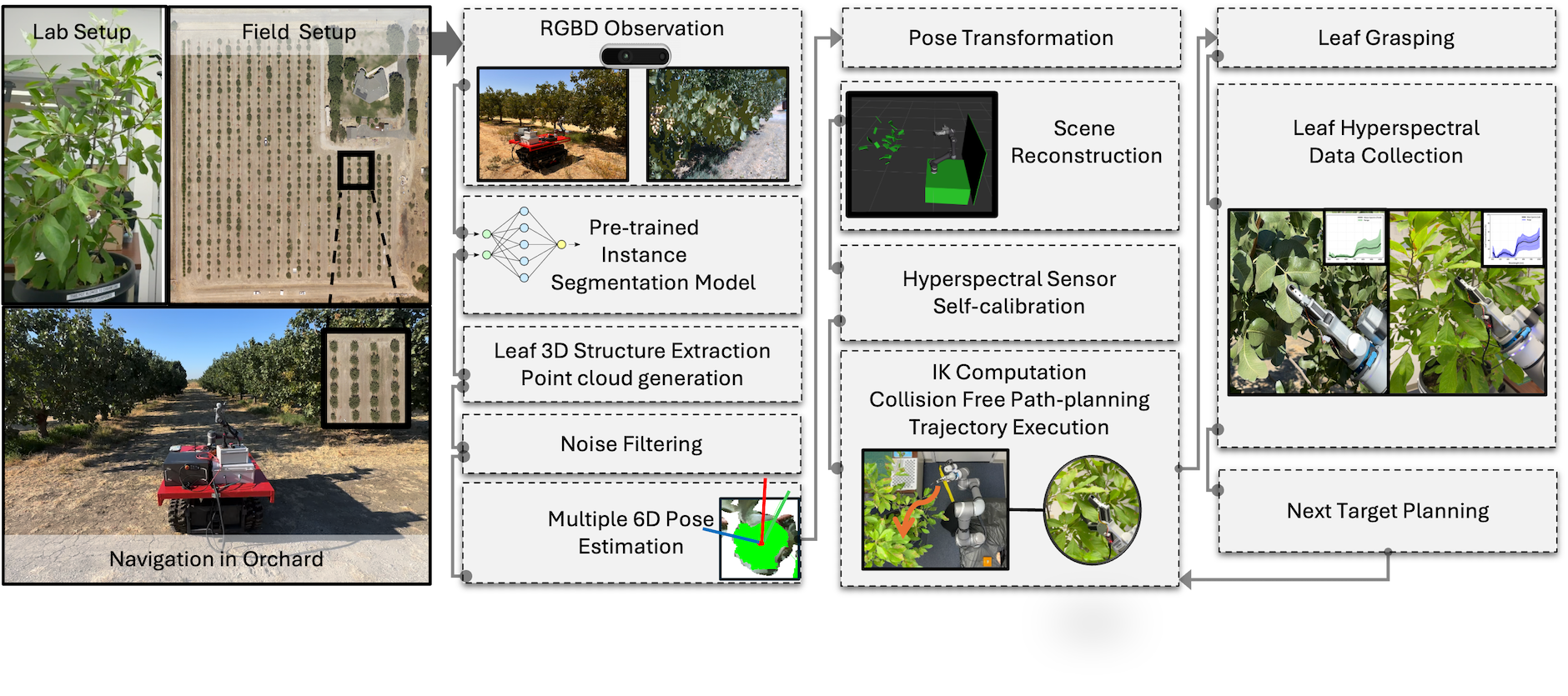}
    \caption{Flowchart for the autonomous proximal hyperspectral leaf sensing process. The workflow consists of a perception pipeline for image processing and 6D pose estimation followed by a robotic manipulation pipeline for precise leaf grasping and hyperspectral data collection.}
    \label{fig:flowchart}
\end{figure*}

In this work, we present RoMu4o, a robotic manipulation unit for orchard operations that autonomously collects hyperspectral data from leaf samples (\cref{fig:system_design}), providing an automated solution for crop health monitoring. Hardware integration benefits from a distinctive compact design that enables high maneuverability for the robotic arm, free from limitations that typically hinder similar systems for hyperspectral imaging. We present a robotic manipulation framework suitable for field-deployable operations. The computer vision pipeline combines a segmentation model trained and fine-tuned for leaf selection, with an algorithmic  procedure for point cloud processing and multi-candidate 6D pose generation. Pose estimates drive a collision-aware motion planner for safe manipulation through dense foliage enabling high-fidelity proximal hyperspectral sampling. We evaluate the system both in the lab on real magnolia plants and in an outdoor pistachio orchard with unstructured, previously unseen foliage; the full pipeline is summarized in \cref{fig:flowchart}.


\section{Related Work}

Most studies in which a robotic manipulation system is developed to physically interact with crops are associated with crop harvesting. Other applications among these systems are extended to tasks such as plant phenotyping, disease detection, trait measurements, physical leaf sampling, and pruning. 

At the intersection of physical interaction with leaves using robotic manipulators, most studies focused on row crops such as maize and sorghum. The Robotanist is among earlier ground robots that was used for manipulation of vegetation such as sorghum and corn \cite{mueller2017robotanist}. A plant phenotyping robotic system using a 4 degree of freedom (DOF) robotic arm and a time-of-flight camera combined with visible and near-infrared (VIS-NIR) spectroscopy and temperature sensors was proposed by \cite{atefi2019vivo}. These experiments were conducted on indoor plant models with grasping success rate of 78\% for maize and 48\% for sorghum. A mobile manipulation robot, named P-AgBot, was developed to perform crop monitoring and physical sampling on corn and sorghum using image processing techniques \cite{kim2022p}. In \cite{deb2023deep}, deep-learning techniques were utilized to provide vision-based guidance to P-AgBot's 6DOF robotic manipulator achieveing 86\% grasp success rate. A proximal hyperspectral imaging system called LeafSpec was integrated with a 6DOF robotic manipulator and tested on greenhouse in-vivo soybean plants achieving a leaf grasping success rate of 93\% \cite{chen2023fully}. A robotic arm was used to reproduce hand-plucking motion for tea leaf harvesting \cite{motokura2020plucking}. An actuation perception framework was developed to physically cut and retrieve leaf samples in which experiments were performed on avocado trees in both indoor and outdoor environments resulting in 69\% success rate in leaf capturing \cite{campbell2022integrated}.

Early studies on leaf manipulation struggled with leaf segmentation and pose estimation within dense foliage. The absence of end-to-end perception-manipulation pipelines further limited real-world performance. Advances in imaging hardware, deep learning–based computer vision models, and robust motion planning algorithms have enabled reliable robotic leaf manipulation in real agricultural environments.

\section{System Overview}
In this section, we explore various components of the robotic hardware and software, outline the experimental setup, and describe the plant models used for spectral measurements in this study. 
\subsection{Hardware Components}
An overview of our integrated hardware is shown in \cref{fig:system_design}. RoMu4o is equipped with a 6-degree-of-freedom collaborative robotic arm (TM5M-900, Techman Robot Inc., Taiwan/OMRON Co., Japan), mounted on a custom-built tank drive ground robot platform. The mobile platform has a dimension of 1.5m x 1.0m x 0.4m (5ft x 3.3ft x 1.2ft) for length, width, and height, and runs on a 24V DC battery. It has four degrees of freedom, one of which is with regard to vertical displacement of the top surface platform. Orchard environments, particularly those featuring pistachio trees, can often reach heights of 6 to 9 meters (20 to 30 feet). To navigate these heights, our platform incorporates the adjustable overhead platform that provides an additional 1-1.5$m$ (up to 5ft) of vertical extension. When combined with the reach of the robotic arm, this configuration grants wider access to the entirety of tree canopy, including regions that are typically beyond the reach of human workers. 

A RealSense D435i RGB-D camera (Intel, USA) and a wide-stroke two-finger gripper (RG2, OnRobot, Denmark) are integrated with a custom-manufactured end-effector designed to encapsulate the optical components. The RealSense depth camera is encapsulated into a lab manufactured casing and installed on top of the built-in RGB camera of the robotic manipulator. The optical sensing component is comprised of a 350-1010nm VIS-NIR hyperspectral spectrometer (NSP32m W1, NanoLambda, South Korea), and an independent lighting source. The light source utilizes 350-2500nm wavelength light for self-calibration and consistent data acquisition regardless of environmental conditions. A VIS-NIR fiber optic cable (F600, StellarNet Inc., USA) is used to transmit light from the source to the end-effector finger gripper. An external computational unit (NUC11TNHi7, Intel, USA) with four 11th generation core processors and 64 GB of RAM is provided for computational processing. We utilized Robot Operating System 2 (ROS2) and MoveIt2 libraries and software to establish communication between sensors and execute the robotic vision and manipulation processes.

\subsection{Experimental Setup}
The leaf manipulation tests were conducted in both indoor and outdoor environments. We selected pistachio tree crops for the outdoor experiments due to their economic significance. A young magnolia shrub with an approximately 2m (5ft) of height is selected as the indoor plant model due to similarities in its leaf structure to pistachios. The test site is an approximately eight-acre pistachio orchard with 11-year-old trees. A small plot within the orchard containing 24 trees was selected to conduct the experiments regarding autonomous hyperspectral data collection from leaf samples. 

The orchard tests were conducted during August 2024, between 8:00-11:30 AM. This timeframe offered optimal sunlight intensity and angle for leaf detection. As the day approached noon, the depth camera performance significantly deteriorated due to overexposure resulting in more missing depth values especially at the leaf boundaries. To mitigate the impact of missing signals and improve the signal-to-noise ratio, we configured the RealSense camera to operate on a medium depth density setting. This setting was found to be an appropriate balance between depth signal density and accuracy in our experimental environments. Additionally, the depth camera was calibrated using a ground truth reference to ensure accurate depth measurements. To prepare the robotic unit for experiments, we tested the system in advance, to identify and address any potential biases arising from misaligned camera installation or reference frame discrepancies. By carefully adjusting for these biases, we ensured that the transformations between the camera's coordinate frame and the robot base frame were accurately aligned so that the gripper can reach exact target locations for leaf grasping with accuracy within one centimeter. 

\section{Leaf grasping pipeline}
\begin{figure*}[t]
    \centering
    \includegraphics[width=1\linewidth]{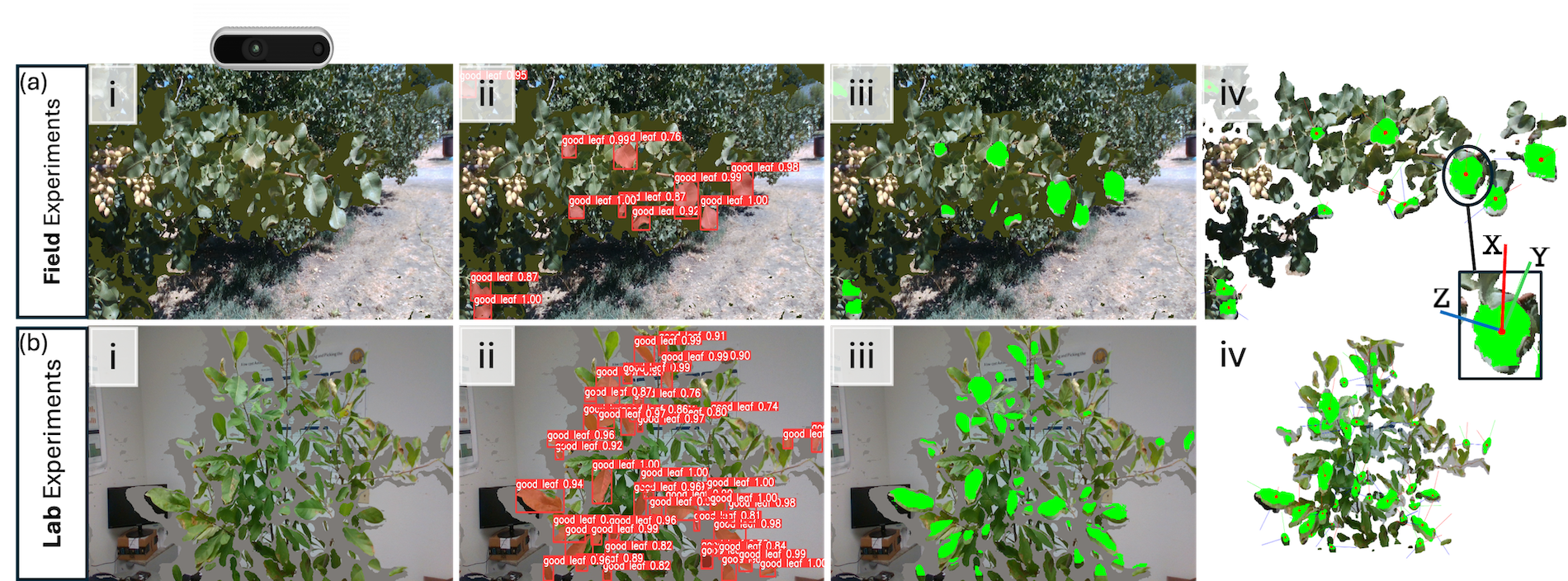}
    \caption{Perception pipeline for leaf manipulation (a) Field and (b) lab experiments showing {(i)} the RGB image as seen by the RealSense depth camera; {(ii)} leaf detection and segmentation using pre-trained instance segmentation deep-learning networks; {(iii)} extracting the 3d structure of identified target leaves by aligning the ordered depth channel to RGB pixels; and {(iv)} noise elimination and 6D pose estimation for each target.}
    \label{fig:vision}
\end{figure*}

A successful grasp in an unstructured environment is the result of a proper perception of the scene complemented by robust motion planning and dexterous grasping. The primary goal of the perception pipeline is to estimate at least one 6D pose for each target leaf within an observed image and identify the collision obstacles occupying the working space of the manipulator that need to be avoided. The systematic procedure demonstrating the perception pipeline is shown in \cref{fig:vision}. The pose information is then processed by the manipulation pipeline to generate joint-space trajectories, which are then executed by the actuators of the robotic arm. In this section, we provide a mathematical framework to explain the details of these two pipelines. A summary of the perception and robotic manipulation pipelines can be seen in \cref{alg:perception} and \cref{alg:manipulation}, respectively.

\subsection{Perception}

\subsubsection{Instance segmentation} 
This pre-processing step leverages a deep learning model not only to distinguish leaves from the background and non-leaf objects but also to identify optimal leaf candidates for robotic manipulation. This dual focus significantly enhances the decision-making process. By training the model on a single category, we detect leaves that are minimally occluded and exhibit ideal visibility and accessibility for robotic grasping. This approach streamlines the leaf selection process within dense foliage, ultimately increasing the chance of successful grasps for spectral sampling. 
There were a total of 872 annotated images, 62\% of which represent the in-lab magnolia leaves and 38\% which represent the in-field pistachio leaves. Two YOLOv8-x instance segmentation models were trained separately to process annotated images for the indoor plant model magnolia leaves and the outdoor crop pistachio leaves with nearly 80\% of the data used for training and the rest split between validation and testing. Results showed a mean average precision of 91\% and 76\% for the in-lab and in-field segmentation models, respectively. During the leaf manipulation experiments, we used a 0.7 confidence threshold to only retain the most reliable detections for further analysis. These models provide the regions of interest (ROIs) within an input image, which are then used to generate 3D point clouds for leaf targets.

\subsubsection{Leaf structure extraction}
Let $\mathcal{O}$ represent an observation by the RGB-D camera, where each observation $ \mathcal{O} \in \mathbb{R}^{H \times W \times 4}$ consists of an image $\mathcal{I} \in \mathbb{R}^{H \times W \times 3}$ and a depth map $\mathcal{D} \in \mathbb{R}^{H \times W}$ with $H$ and $W$ denoting the image height and width, respectively. An instance segmentation model $\mathcal{S}: \mathbb{R}^{H \times W \times 3} \to \mathbb{B}^{H \times W \times N}$ with  $\mathbb{B} = \{0,1\}$, processes the RGB image $\mathcal{I}$ to produce $N$ binary masks $\mathcal{M} = \{ {M}_j \mid {M}_j \in \mathbb{B}^{H \times W} \, , \, j=1,...,N\}$, where ${M}_j $ corresponds to individual leaf masks detected in the image $\mathcal{I}$ . For each mask ${M}_j$, we extract the corresponding depth values from $\mathcal{D}$ to form masked depth maps ${D}_j = \mathcal{D} \odot {M}_j $ containing the depth values of each identified leaf in $\mathcal{O}$ considering ${ D} \in \mathbb{R}^{H \times W}$. 

\subsubsection{Point cloud reconstruction}

Using the intrinsic parameters of the camera $(f_x,f_y , c_x , c_y)$ and the operator $\mathcal{T}(D_j(u,v), f_x,f_y , c_x , c_y )$ where $\mathcal{T} : \mathbb{R}^{H \times W} \to \mathbb{R}^{H \times W  \times 3} $, we transform the extracted depth maps $D_j(u,v)$  into ${P}_j(u,v)$ to represent the individual 3D point clouds for each identified leaf while $u$  and $v$ show the pixel locations. Note that $P_j$ contains non-zero values for masked pixels where a leaf is identified and zero values anywhere else. Let $\mathcal{P}_j  \in \mathbb{R}^{m_j \times 3}$ represent a new set that contains only the 3D point clouds of each leaf with $m$ denoting the number of available points for the $j^{ th} $ leaf instance. 

We filter out potential noise from each individual point cloud $\mathcal{P}_j$ to ensure only valid points are retained for subsequent analysis. The noise filtering process aims to remove erroneous points and outliers that misrepresent the surface of the desired leaf. These unwanted clusters can arise from several factors, such as segmentation errors containing background pixels or occlusions from overlapping leaves. We adopted a Gaussian-based outlier rejection method to address the potential anomaly in the initial extracted point cloud. Consider ${\mathcal{P}}_{j}^k \sim \mathcal{N}({\mu}_j^k, {\sigma^2}^k_j)$ with $k=1,2,3$ and ${\mu}_j^k$ and $ {\sigma^2}^{k}_j$ denoting the mean and standard deviation across each of the three-dimensional axes. The z-scores for each point in \(\mathcal{P}^k_j\) are then calculated. For each dimension \(k\), the z-score for points in \(\mathcal{P}_j^k\) is calculated by $Z_j^k = (\mathcal{P}_j^k - \mu_j^k) / {\sigma_j^k}$ such that $\mathcal{P}_j^k , Z^k_j \in \mathbb{R}^{m_j}$ and ${\mu}_j^k , \sigma^k_j \in \mathbb{R}$. To filter out noise, we establish a threshold of $Z_{th} = 2.33$ which corresponds to a confidence interval of $98\%$. Thus, any points with a z-score greater than \(Z_{\text{th}}\) in any dimension are considered outliers and removed. The filtered point cloud $\hat{\mathcal{P}}_j$ is then constructed by retaining the points that satisfy $|Z_{j,i}^k| \leq Z_{th} \,\, \forall i=1,2,...,m_j$. 

\subsubsection{Pose estimation}  We estimate the central point for each filtered point cloud $\hat{\mathcal{P}}_j$ to localize the target leaves with respect to the camera reference frame. Let $\hat{\mathcal{P}}_j = \{ {\hat{\mathbf{p}}}_{j,1},..., {\hat{\mathbf{p}}}_{j, m_j} \}$ such that $\hat{\mathbf{p}}_{j, i} \in \mathbb{R}^3$ and define the central point ${\overline{\mathbf{p}}}_{j} \in \hat{\mathcal{P}}_j$ which will be found by computing the closest point to the median ${\Tilde{\mathbf{p}}}_{j}$. Therefore, the central point ${\overline{\mathbf{p}}}_{j}$ can be written by:

\begin{equation}
{\overline{\mathbf{p}}}_{j} = \arg\min_{\mathbf{x} \in \hat{\mathcal{P}}_j} \left\| \mathbf{x} - {\Tilde{\mathbf{p}}}_{j} \right\|_2
\end{equation}

To determine the orientation of each leaf, we perform principal component analysis (PCA) on each point cloud. Using PCA, we find the three principal axes of variance in $\hat{\mathcal{P}}_j$.  A singular value decomposition technique is used to decompose the covariance matrix of points and provide eigenvalues and eigenvectors associated with the three orthogonal vectors that best explain the variance through the data. The eigenvalues $\lambda_1, \lambda_2, \lambda_3$  represent the magnitude of variance along the eigenvectors \(\boldsymbol{\nu}_1, \boldsymbol{\nu}_2, \boldsymbol{\nu}_3\). The eigenvector corresponding to the smallest eigenvalue aligns with the direction of least variance in the point cloud. This unique eigenvector is perpendicular to the plane that best fits the surface of the leaf. We denote this unit vector as ${\mathbf{n}}_j$ representing the normal vector to the surface of leaf $j$. The normal vectors are adjusted to ensure they are oriented toward the viewpoint, with their direction reversed if initially oriented away from the viewpoint. 

\begin{algorithm}[h]
\caption{Perception Pipeline}
\label{alg:perception}
\begin{algorithmic}[1]
\renewcommand{\algorithmicrequire}{\textbf{Input:}}
\renewcommand{\algorithmicensure}{\textbf{Output:}}
\REQUIRE RGBD observation:  $\mathcal{O} = \{\mathcal{I}, \mathcal{D}\}$
\ENSURE Leaf pose estimates:  $\mathcal{L}_j^{\{C\}}$

\STATE $\mathcal{M} \leftarrow \mathcal{S}(\mathcal{I})$

\FOR{number of masks $j = 1$ to $N$}
    \STATE $D_j \leftarrow \mathcal{D} \odot M_j$
    \STATE $\mathcal{P}_j \leftarrow \mathcal{T}(D_j(u, v), f_x, f_y, c_x, c_y) $
    
    \FOR{$k = 1,2,3$}
        \STATE $\mu_j^k, {\sigma^2}^k_j \leftarrow (\mathcal{P}_j^k)$
        \STATE $Z_j^k \leftarrow  \mathcal{N}({\mu}_j^k, {\sigma^2}^k_j)$
        \FOR{number of points $i = 1$ to $m_j$}
           \STATE $\hat{\mathcal{P}}_j \leftarrow \{\mathbf{p} \in \mathcal{P}_j \, \, \, \mathbf{if}  \,\,\, |Z_{j,i}^k| \leq Z_{th}\}$
        \ENDFOR
    \ENDFOR

    \STATE ${\overline{\mathbf{p}}}_{j} \leftarrow \arg\min_{\mathbf{x} \in \hat{\mathcal{P}}_j} \left\| \mathbf{x} - {\Tilde{\mathbf{p}}}_{j} \right\|_2$
    \STATE $\{\lambda_k, \boldsymbol{\nu}_k\}_{k=1}^3 \leftarrow \text{PCA}(\hat{\mathcal{P}}_j)$
    \STATE $\mathbf{n}_j \leftarrow {\arg\min}_{\boldsymbol{\nu}_k} \, \, \lambda_k$  
    \STATE $\mathbf{t}_j \leftarrow \mathbf{p}^*_j - {\overline{\mathbf{p}}}_{j}$
    \STATE $\mathbf{b}_j \leftarrow \mathbf{n}_j \times \mathbf{t}_j$
    \STATE $(\overline{\mathbf{p}}_j, \mathbf{q}_{j,1}) \leftarrow (\overline{\mathbf{p}}_j, \mathbf{t}_j,\mathbf{b}_j,\mathbf{n}_j) \in SE(3) $ 
    \STATE $\mathcal{L}_j^{\{C\}} \leftarrow  (\overline{\mathbf{p}}_j, \mathbf{q}_{j,1},  \alpha )$
\ENDFOR

\end{algorithmic} 
\end{algorithm}

To establish an orthogonal local coordinate system on each leaf with unit vectors $({\mathbf{t}}_j, {\mathbf{b}}_j, {\mathbf{n}}_j)$, we need to determine ${\mathbf{t}}_j$ and ${\mathbf{b}}_j$ such that they lie within the leaf surface plane. To determine ${\mathbf{t}}_j$  we first select a reference point on the edge of the point cloud that represents the uppermost point along the vertical axis of the camera. This point provides an initial guess close to the stem attachment point. Let $\mathbf{v}^*_j =  \mathbf{p}^*_j - {\overline{\mathbf{p}}}_{j} $ serve as a vector that connects this point to the center of the leaf. To ensure $\mathbf{v}^{*}_j$ lies within the surface plane, we project it onto the plane defined by ${\mathbf{n}}_j$. This projection can be written as ${\mathbf{v}}_j = \mathbf{v}^*_j - (\mathbf{v}^*_j \cdot {\mathbf{n}}_j) {\mathbf{n}}_j$ so that we arrive at ${\mathbf{t}}_j = \mathbf{v}_j/\left\| \mathbf{v}_j \right\|$. With ${\mathbf{t}}_j$ and ${\mathbf{n}}_j$ available we compute their cross product to obtain \({\mathbf{b}}_j = {\mathbf{n}}_j \times {\mathbf{t}}_j\). We obtain a 6D pose for each leaf by establishing an orthogonal coordinate system centered at ${\overline{\mathbf{p}}}_{j}$ with orientations given by unit vectors ${\mathbf{t}}_j$, ${\mathbf{b}}_j$, and ${\mathbf{n}}_j$.

Given the fact that our ultimate goal is to collect leaf hyperspectral data, we are not limited to a unique 6D pose. Rather we aim to propose a solution that increases the chance of successful leaf grasping by the robotic manipulator. Thus, we propose a total of five 6D poses for each target leaf. The additional four poses are derived from the initial guess with four counter clock wise $\alpha = \{ -\pi/4 ,-\pi/2, -3\pi/4 , \pi \}$ intrinsic rotations along the normal vector $\mathbf{n}_j$. Lastly, these rotations will be computed and the resulting 6D poses for each leaf are obtained. Let  $\mathcal{L}_j^{ \{C\}}= \{(\overline{\mathbf{p}}_{j}, \mathbf{q}_{j,1}), ... , (\{\overline{\mathbf{p}}_{j}, \mathbf{q}_{j,5})\}$ where $\mathbf{q} $ denote the rotation quaternions and $\{C\}$ represent the camera reference frame.  

\subsection{Robotic Manipulation}
Given the obtained pose information, we construct a robotic manipulation workflow for effective leaf grasping and hyperspectral data sampling using the integrated optical sensor. This workflow establishes a collision-aware motion planning framework that accounts for both the kinematic and task-specific constraints. Leaf 6D poses are transformed from the camera reference frame $\mathcal{L}_j^{\{C\}}$ into the robot base frame $\mathcal{L}_j^{\{B\}}$. 

\begin{algorithm}[h]
 \caption{Robotic Manipulation Workflow}
 \label{alg:manipulation}
 \begin{algorithmic}[1]
 \renewcommand{\algorithmicrequire}{\textbf{Input:}}
 \renewcommand{\algorithmicensure}{\textbf{Output:}}
 \REQUIRE  \(\mathcal{L}_j^{\{C\}}\)
 \ENSURE Leaf hyperspectral data

 \STATE   \(\mathcal{L}_j^{\{B\}}\) $\leftarrow $ \( \mathcal{L}_j^{\{C\}}\) 
 \STATE Add collision objects to scene
 \STATE self-calibrate hyperspectral sensor
 \FOR{number of leaf targets: $j = 1$ to $N$}
 \FOR{each pose in \(\mathcal{L}_{j}^{\{B\}}: \,  n=1 \) to $5$ } 
    \STATE Run IK Solver and RRTC $\mid$ $\mathcal{G}^{\{B\}} := \mathcal{L}_{j}^{\{B\}}$
    \IF{solution exists}
       \STATE Compute and execute trajectory
       \STATE Control actuation and verify goal reached
       \STATE Trigger gripper to capture leaf 
       \STATE Acquire hyperspectral data
       \STATE Break; proceed to leaf $j+1$
    \ELSE
       \STATE Continue to pose \( n+1 \) 
    \ENDIF
 \ENDFOR
 \ENDFOR
 \STATE Save collected data
 \STATE Return gripper to initial state
 \STATE Go to next batch with new observation:  $\mathcal{O}^{new} = \{\mathcal{I}, \mathcal{D}\}$
 \STATE Run perception and manipulation pipelines
 \end{algorithmic} 
\end{algorithm}

The planning scene is populated with geometrically simplified representations of each leaf to prevent unintended collisions as a contact may result in missed targets or compromised spectral data. The objective is then to align the gripper reference frame $G^{\{B\}}$ with the identified target poses $\mathcal{L}_j^{\{B\}}$ while adhering to kinematic constraints and minimizing collision risk. This alignment relies on an inverse kinematics (IK) solver. We use a numerical, Jacobian-based solver to compute valid joint configurations for the robotic arm to reach the desired end-effector poses. Once a feasible joint-space solution is obtained through the IK solver, a rapidly-exploring random tree connect (RRTC) path planner is employed to generate a collision-free path from the current state to the target state of the robotic arm. The RRTC path planner operates by concurrently building two rapidly exploring random trees, one rooted at the start state and the other at the goal state within the configuration space. The trees expand by iteratively sampling random states, steering toward the sampled points and connecting them through feasible joint space while avoiding collisions with both the robot itself as well as the predefined objects. Once a connection between the trees is established, a collision-free path is generated. Our planner leverages parallel exploration of the configuration space to find optimized paths. Two design choices are specific to leaf grasping rather than generic manipulation: the planning scene is populated with per-leaf collision primitives so that the arm treats surrounding foliage as obstacles to preserve uncaptured leaves and avoid disturbing the target, and the five candidate poses per leaf are attempted in sequence so that a single unreachable approach does not forfeit the target. According to the obtained paths, the manipulation workflow first generates and executes a trajectory to the initial target, then sequentially transitions into processing solutions to the subsequent targets as each task is completed. During trajectory execution, motor controllers continuously monitor the joint states and compare their real-time poses with the target state. Adjustments to the actuation are made as needed to maintain precise alignment with the planned trajectory. Upon reaching each target, the gripper captures the target leaf and allows the built-in optical sensor system to acquire the hyperspectral data.

\begin{figure*}[t]
    \centering
    \includegraphics[width=1\linewidth]{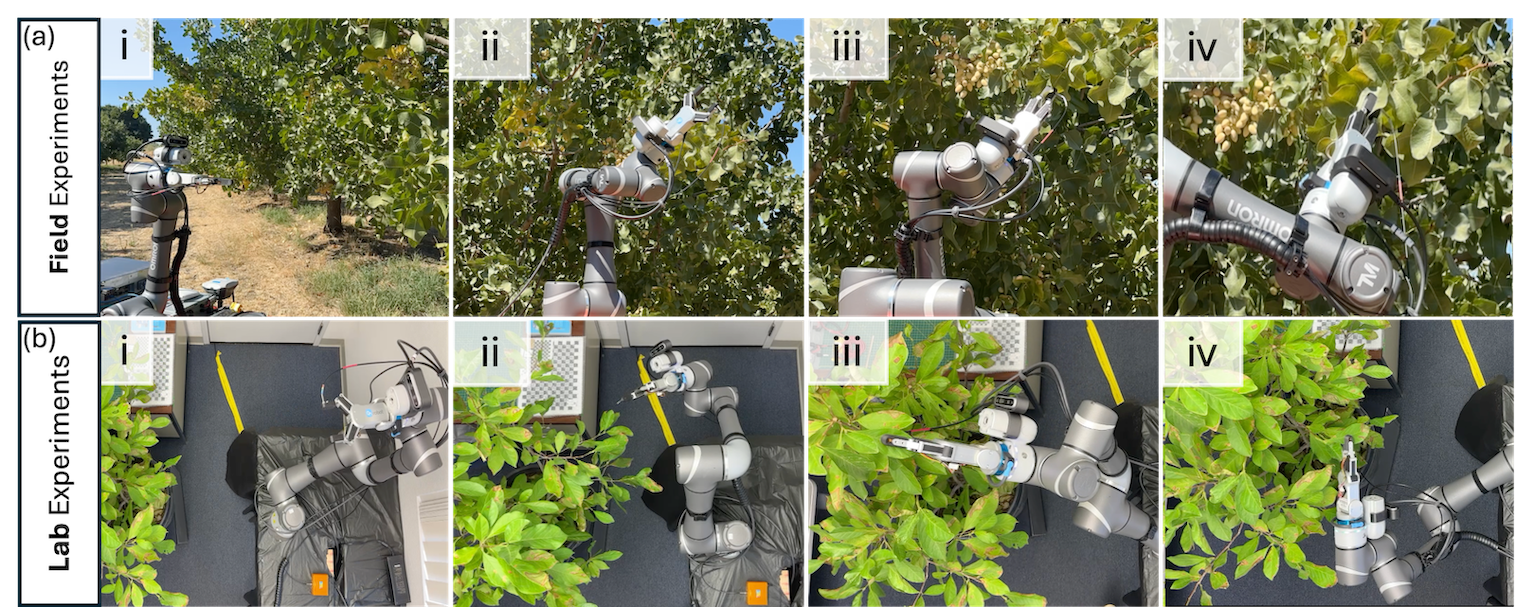}
    \caption{Leaf manipulation process in action. (a) Pistachio trees representing field experiments on the \textit{top row}; (b) magnolia shrub representing lab experiments on the \textit{bottom row}. {(i)} a batch of foliage is observed by the camera for the perception pipeline to generate 6D poses for leaf targets; {(ii)} pose estimates are processed through the robotic manipulation workflow for path planning and trajectory execution while the hyperspectral sensor is calibrated; {(iii)} the first target is approached and grasped, hyperspectral data collected and stored; {(iv)} planning to the next target for leaf grasping and spectroscopy measurements.} 
    \label{fig:manipulation}
\end{figure*}

\section{Results and Analysis}
This robotic unit is designed to operate in orchards and autonomously collect hyperspectral data from leaf samples. In this work, our design was primarily focused on pistachio trees. However, to facilitate system development and refinement, we initially conducted experiments indoors using magnolia plant models due to the similarity in leaf structure between magnolia and pistachio leaves. This approach allowed us to iteratively test and enhance the robotic system in a controlled environment before field trials. 

The perception-manipulation pipeline is executed sequentially to identify and grasp leaf targets. A demonstration of the leaf manipulation process in action is shown in \cref{fig:manipulation}. To begin each experiment the robotic arm is positioned approximately 50$cm$ away from a target \textit{batch} of the foliage. The perception pipeline is then activated to analyze the scene, identify leaf samples that are suitable for manipulation, extract their 3D structure, and propose 6D poses that are positioned at the center of leaf surfaces. These 6D poses are then organized with respect to their distance to the camera reference frame and transferred into the robotic manipulation workflow for grasping and hyperspectral data sampling. While multiple 6D poses are identified within each batch, the manipulation pipeline may not always execute a trajectory for the robotic arm to approach all targets. This can be due to certain poses being highly clustered or out of reach resulting in an unavailable path or IK solution. An \textit{approach} is considered when the manipulation pipeline is able to plan and execute a trajectory for the robotic arm to approach a target leaf, regardless of the outcome. If an approach results in successful leaf grasping and hyperspectral data collection, this is considered a \textit{successful approach}.

\begin{figure*}[ht]
    \centering
    \includegraphics[width=.78\linewidth]{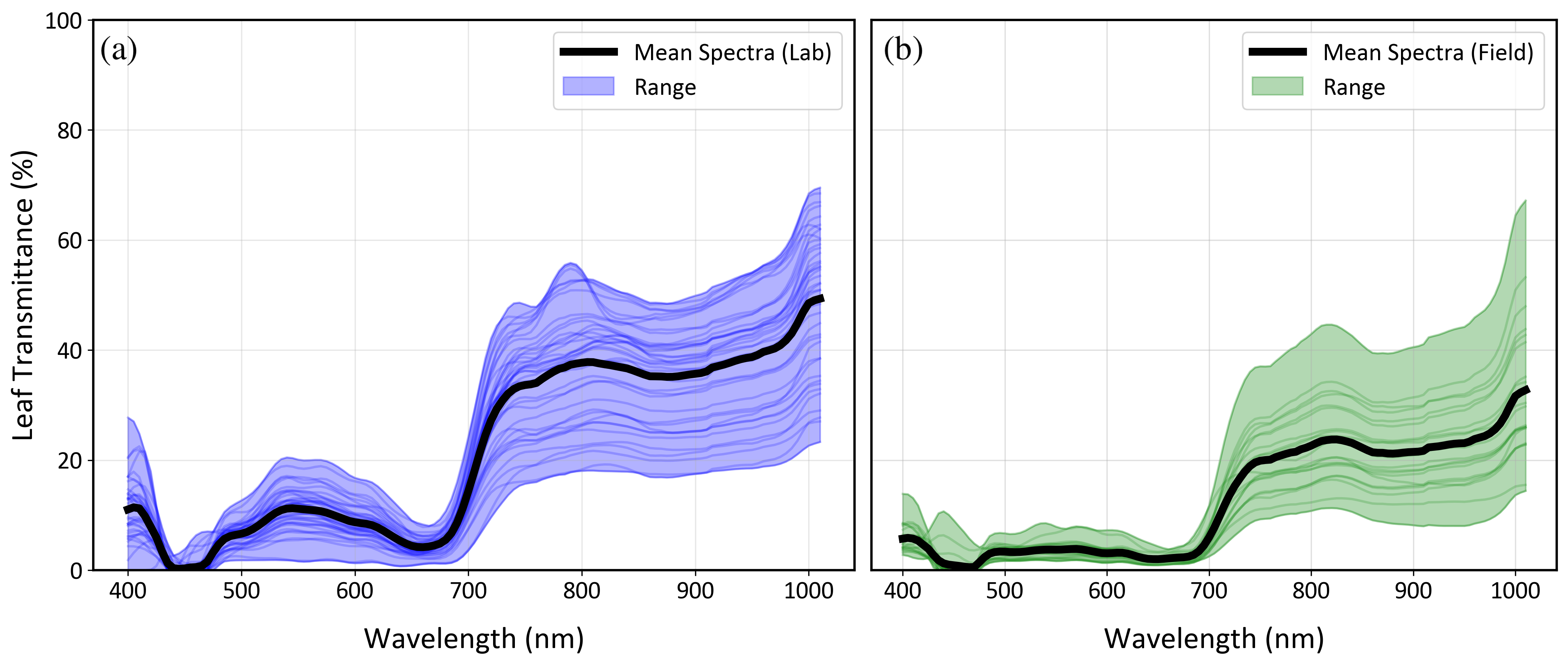}
    \caption{Leaf hyperspectral data collected using our proposed robotic system; ({a}) Spectral transmittance of magnolia leaves used as a plant model for indoor lab experiments; ({b}) Spectral transmittance of pistachio leaves representing our outdoor field experiments. Note that all hyperspectral data is baseline corrected.}
    \label{fig:leaf_spectra}
\end{figure*}

\subsection{Leaf Manipulation Performance}
The leaf grasping success rate considering the total approaches is provided in \cref{tab:total_approaches}. In total, there were 131 approaches by the robotic arm resulting in leaf grasping and data sampling. Of these, 75 approaches were performed on magnolia leaves as the indoor plant model, while 56 approaches were conducted on pistachio leaves during outdoor field experiments. For in-lab and in-field trials, the overall success rates considering all approaches were 63\% and 70\%, respectively. Considering both settings combined, 66\% success rate was achieved. 

\begin{table}[htb]
\centering
\caption{Success rate of leaf grasping using the robotic manipulator in different experimental settings.}
\renewcommand{\arraystretch}{1.3} 
\small
\begin{tabular}{c p{2.5cm} p{2cm}} 
\midrule
\bottomrule
\hline
 & \centering\shortstack{Total \\ Approaches} & \shortstack{\\ Grasping \\ Success \\ Rate (\%)}  \\ \midrule
In-Lab    & \centering 75 & \hspace{5mm}63  \\ \hline
In-Field  & \centering 56 & \hspace{5mm}\textbf{70}   \\ \hline
Combined  & \centering 131 & \hspace{5mm}66  \\ \bottomrule 
\end{tabular}
\label{tab:total_approaches}
\end{table}

\subsection{LPB Performance}
We use LPB to denote leaf per batch when further analyzing successful performance of RoMu4o. All leaf batches contain one or more approaches. Reporting LPB is motivated by the real application of proximal hyperspectral leaf sensing in agricultural environments. Researchers and field technicians commonly use spectroradiometers to manually measure data from one or more leaf samples per tree. We report LPB metrics as a direct measure of RoMu4o’s autonomous operational performance for up to three leaves per batch. Therefore, we can denote batches that contain at least 1 leaf approached by the robotic manipulator as 1-LPB.  If all or some of these batches had at least 2 leaves approached by the robotic manipulator, they will be denoted as 2-LPB. Similarly, if they had at least 3 leaves approached, they are denoted as 3-LPB.  The percent of batches in each of these scenarios is shown in Table~\ref{tab:available_tests}.

\begin{table}[htb]
\centering
\caption{Percentage of batches that contain at least 1, 2 or 3 leaves approached.}
\renewcommand{\arraystretch}{1.3} 
\small
\begin{tabular}{c p{1.3cm} p{1.3cm} p{1.3cm} p{1.3cm}} 
\midrule
\bottomrule
\hline
 & \shortstack{\\ 1-LPB \\ (\%)} &  \shortstack{ \\  2-LPB \\ (\%) } & \shortstack{ \\ 3-LPB \\ (\%) } \\ \midrule
In-Lab    &  \hspace{3mm}100 & \hspace{3mm}100 &\hspace{3mm}77  \\ \hline
In-Field  &  \hspace{3mm}100 & \hspace{4mm}71 &\hspace{3mm}42 \\ \hline
Combined  &  \hspace{3mm}100 & \hspace{4mm}85 &\hspace{3mm}59 \\ \bottomrule 

\end{tabular}
\label{tab:available_tests}
\end{table}

In-lab tests were comprised of 22 total batches, all of which contained at least 2 leaves approached per batch, and 17 of which contained 3-LPB. These values represent 100\% for both 1- and 2-LPB, and 77\% for 3-LPB. Field experiments consisted of a total of 24 batches, with 17 and 10 of them containing 2-LPB and 3-LPB, respectively. These values represent 100\%, 71\%, and 42\% for 1-, 2-, and 3-LPB, respectively. In both indoor and outdoor experiments combined, there were 46, 39, and 27 batches with 1-, 2-, and 3-LPB, respectively. These values represent 100\% for 1-LPB followed by 85\% for 2-LPB and 59\% for 3-LPB.

\begin{table}[h]
\centering
\caption{Success rate of leaf grasping per LPB group.}
\renewcommand{\arraystretch}{1.3} 
\small
\begin{tabular}{c p{1.3cm} p{1.3cm} p{1.3cm} p{1.3cm}} 
\midrule
\bottomrule
\hline
  & \shortstack{\\ 1-LPB \\ (\%)} &  \shortstack{ \\  2-LPB \\ (\%) } & \shortstack{ \\ 3-LPB \\ (\%) } \\ \midrule
In-Lab     &  \hspace{3mm}95 & \hspace{3mm}77 &\hspace{3mm}41  \\ \hline
In-Field   &  \hspace{3mm}79 & \hspace{3mm}70 &\hspace{3mm}60 \\ \hline
Combined   &  \hspace{3mm}87 & \hspace{3mm}74 &\hspace{3mm}48 \\ \bottomrule

\end{tabular}
\label{tab:test_results}
\end{table}

In \cref{tab:test_results}, the success rate of leaf grasping per batch is reported.
Leaf manipulation experiments on the indoor plant model, yielded 95\% for 1-LPB, dropping to 77\% for 2-LPB and 41\% for 3-LPB. For example, the value for 2-LPB shows that 77\% of the batches in 2-LPB group contained at least 2 successful grasps. In field experiments, the success rate for the 1-LPB group was 79\% which was followed by 70\% and 60\% for 2- and 3-LPB, respectively. Both indoor and outdoor results combined reflected 87\% success rate for 1-LPB group, with a gradual decrease to 74\% for 2-LPB followed by 48\% for 3-LPB. For these successful grasps, leaf hyperspectral data on a 400-1010$nm$ range is collected and the results are demonstrated in \cref{fig:leaf_spectra}. All hyperspectral data in this figure have been baseline corrected using an initial self-calibration procedure that records the maximum light intensity in the absence of a leaf. These high-fidelity hyperspectral data can be used for phenotyping purposes, early detection of biotic or abiotic stress, and nutrient deficiencies.

\section{Conclusions}
We presented RoMu4o, a robotic manipulation unit for orchard operations, designed to address the need for continuous and automated orchard monitoring. RoMu4o is engineered for proximal hyperspectral leaf sensing and leverages the compact integration of a hyperspectral spectrometer and an independent lighting source with the end-effector of a 6-DOF robotic arm. We proposed an algorithmic procedure for image processing and 6D pose estimation. Our computer vision method relies on a deep-learning instance segmentation model trained to identify minimally occluded leaf candidates deemed suitable for robotic manipulation. Furthermore, we created a collision-aware motion planning workflow for robust robotic manipulation and leaf grasping. System performance was evaluated through experiments on real magnolia plants in the lab setting, followed by field trials in a pistachio orchard with unstructured, previously unseen batches of foliage. The system demonstrated reliable performance for 1-LPB hyperspectral sampling under both controlled and uncontrolled conditions, achieving a 95\% success rate in lab trials and 79\% in field trials. Overall, the field experiments yielded a grasping success rate of 70\%, which highlights the ability and performance of our robotic unit in unstructured orchard environments.

RoMu4o demonstrates the feasibility of robotic leaf manipulation in real orchards, though limitations remain. Outdoor operations suffered from intense sunlight, which degrades the camera's depth estimation at leaf boundaries. Electronic hardware lost performance as ambient temperature increased. Furthermore, our crop-specific segmentation models leave cross-species generalization and long-term throughput untested. Field experiments in real orchards carry high operational costs, restricting the number of trials that can be performed within the crop's phenological growing cycle. This opens an opportunity for complementary work using RoMu4o, or similar variants, to mitigate sample-size constraints. Other promising directions include utilization of multi-modal sensor fusion and closed-loop visual servoing to compensate for potential leaf displacements and highly curved leaf targets. Replacing CNN-based perception and classical planners with transformer-, VLA-, or WMA-based architectures may further improve throughput, cycle-time and scalability. However, whether such foundational models can justify their computational costs on edge hardware, maintain low latency during the operation, and satisfy targeted evaluation metrics for safe field deployments remains an open question.

\bibliography{references}

@INPROCEEDINGS{Kim2024deformable,
  author={Kim, Chung Hee and Lee, Moonyoung and Kroemer, Oliver and Kantor, George},
  booktitle={2024 IEEE International Conference on Robotics and Automation (ICRA)}, 
  title={Towards Robotic Tree Manipulation: Leveraging Graph Representations}, 
  year={2024},
  volume={},
  number={},
  pages={11884-11890},
  doi={10.1109/ICRA57147.2024.10611327}}

@article{lee2024towards,
  title={Towards autonomous crop monitoring: Inserting sensors in cluttered environments},
  author={Lee, Moonyoung and Berger, Aaron and Guri, Dominic and Zhang, Kevin and Coffey, Lisa and Kantor, George and Kroemer, Oliver},
  journal={IEEE Robotics and Automation Letters},
  volume={9},
  number={6},
  pages={5150--5157},
  year={2024},
  publisher={IEEE}
}

@article{kim2025autonomous,
  title={Autonomous robotic pepper harvesting: Imitation learning in unstructured agricultural environments},
  author={Kim, Chung Hee and Silwal, Abhisesh and Kantor, George},
  journal={IEEE Robotics and Automation Letters},
  year={2025},
  publisher={IEEE}
}

@article{motokura2020plucking,
  title={Plucking motions for tea harvesting robots using probabilistic movement primitives},
  author={Motokura, Kurena and Takahashi, Masaki and Ewerton, Marco and Peters, Jan},
  journal={IEEE Robotics and Automation Letters},
  volume={5},
  number={2},
  pages={3275--3282},
  year={2020},
  publisher={IEEE}
}

@article{junttila2022close,
  title={Close-range hyperspectral spectroscopy reveals leaf water content dynamics},
  author={Junttila, Samuli and H{\"o}ltt{\"a}, Tiina and Saarinen, Ninni and Kankare, Ville and Yrttimaa, Tuomas and Hyypp{\"a}, Juha and Vastaranta, Mikko},
  journal={Remote Sensing of Environment},
  volume={277},
  pages={113071},
  year={2022},
  publisher={Elsevier}
}

@article{you2023semiautonomous,
  title={Semiautonomous precision pruning of upright fruiting offshoot orchard systems: An integrated approach},
  author={You, Alexander and Parayil, Nidhi and Krishna, Josyula Gopala and Bhattarai, Uddhav and Sapkota, Ranjan and Ahmed, Dawood and Whiting, Matthew and Karkee, Manoj and Grimm, Cindy M and Davidson, Joseph R},
  journal={IEEE Robotics \& Automation Magazine},
  year={2023},
  publisher={IEEE}
}

@article{xu2022review,
  title={A review of high-throughput field phenotyping systems: focusing on ground robots},
  author={Xu, Rui and Li, Changying},
  journal={Plant Phenomics},
  year={2022},
  publisher={AAAS}
}

@inproceedings{mueller2017robotanist,
  title={The Robotanist: A ground-based agricultural robot for high-throughput crop phenotyping},
  author={Mueller-Sim, Tim and Jenkins, Merritt and Abel, Justin and Kantor, George},
  booktitle={2017 IEEE international conference on robotics and automation (ICRA)},
  pages={3634--3639},
  year={2017},
  organization={IEEE}
}

@article{chen2023fully,
  title={Fully automated proximal hyperspectral imaging system for high-resolution and high-quality in vivo soybean phenotyping},
  author={Chen, Ziling and Wang, Jialei and Jin, Jian},
  journal={Precision Agriculture},
  volume={24},
  number={6},
  pages={2395--2415},
  year={2023},
  publisher={Springer}
}

@inproceedings{deb2023deep,
  title={Deep Learning-Based Leaf Detection for Robotic Physical Sampling with P-AgBot},
  author={Deb, Aarya and Kim, Kitae and Cappelleri, David J},
  booktitle={2023 IEEE/RSJ International Conference on Intelligent Robots and Systems (IROS)},
  pages={8291--8297},
  year={2023},
  organization={IEEE}
}

@inproceedings{campbell2022integrated,
  title={An integrated actuation-perception framework for robotic leaf retrieval: detection, localization, and cutting},
  author={Campbell, Merrick and Dechemi, Amel and Karydis, Konstantinos},
  booktitle={2022 IEEE/RSJ International Conference on Intelligent Robots and Systems (IROS)},
  pages={9210--9216},
  year={2022},
  organization={IEEE}
}

@article{atefi2019vivo,
  title={In vivo human-like robotic phenotyping of leaf traits in maize and sorghum in greenhouse},
  author={Atefi, Abbas and Ge, Yufeng and Pitla, Santosh and Schnable, James},
  journal={Computers and Electronics in Agriculture},
  volume={163},
  pages={104854},
  year={2019},
  publisher={Elsevier}
}

@article{lei20244d,
  title={4D Metric-Semantic Mapping for Persistent Orchard Monitoring: Method and Dataset},
  author={Lei, Jiuzhou and Prabhu, Ankit and Liu, Xu and Cladera, Fernando and Mortazavi, Mehrad and Ehsani, Reza and Chaudhari, Pratik and Kumar, Vijay},
  journal={arXiv preprint arXiv:2409.19786},
  year={2024}
}

@article{yoshida2022fruit,
  title={Fruit recognition method for a harvesting robot with RGB-D cameras},
  author={Yoshida, Takeshi and Kawahara, Takuya and Fukao, Takanori},
  journal={ROBOMECH Journal},
  volume={9},
  number={1},
  pages={1--10},
  year={2022},
  publisher={Springer}
}

@article{kim2022p,
  title={P-AgBot: In-Row \& Under-Canopy Agricultural Robot for Monitoring and Physical Sampling},
  author={Kim, Kitae and Deb, Aarya and Cappelleri, David J},
  journal={IEEE Robotics and Automation Letters},
  volume={7},
  number={3},
  pages={7942--7949},
  year={2022},
  publisher={IEEE}
}

@article{zhou2022intelligent,
  title={Intelligent robots for fruit harvesting: Recent developments and future challenges},
  author={Zhou, Hongyu and Wang, Xing and Au, Wesley and Kang, Hanwen and Chen, Chao},
  journal={Precision Agriculture},
  pages={1--52},
  year={2022},
  publisher={Springer}
}

@article{wang2021smart,
  title={From smart farming towards unmanned farms: A new mode of agricultural production},
  author={Wang, Tan and Xu, Xianbao and Wang, Cong and Li, Zhen and Li, Daoliang},
  journal={Agriculture},
  volume={11},
  number={2},
  pages={145},
  year={2021},
  publisher={MDPI}
}

@article{gao2018novel,
  title={A novel multirobot system for plant phenotyping},
  author={Gao, Tianshuang and Emadi, Hamid and Saha, Homagni and Zhang, Jiaoping and Lofquist, Alec and Singh, Arti and Ganapathysubramanian, Baskar and Sarkar, Soumik and Singh, Asheesh K and Bhattacharya, Sourabh},
  journal={Robotics},
  volume={7},
  number={4},
  pages={61},
  year={2018},
  publisher={MDPI}
}

@inproceedings{thayer2018multi,
  title={Multi-robot routing algorithms for robots operating in vineyards},
  author={Thayer, Thomas C and Vougioukas, Stavros and Goldberg, Ken and Carpin, Stefano},
  booktitle={2018 IEEE 14th International Conference on Automation Science and Engineering (CASE)},
  pages={14--21},
  year={2018},
  organization={IEEE}
}

@inproceedings{albani2017monitoring,
  title={Monitoring and mapping with robot swarms for agricultural applications},
  author={Albani, Dario and IJsselmuiden, Joris and Haken, Ramon and Trianni, Vito},
  booktitle={2017 14th IEEE International Conference on Advanced Video and Signal Based Surveillance (AVSS)},
  pages={1--6},
  year={2017},
  organization={IEEE}
}

@inproceedings{blender2016managing,
  title={Managing a mobile agricultural robot swarm for a seeding task},
  author={Blender, Timo and Buchner, Thiemo and Fernandez, Benjamin and Pichlmaier, Benno and Schlegel, Christian},
  booktitle={IECON 2016-42nd Annual Conference of the IEEE Industrial Electronics Society},
  pages={6879--6886},
  year={2016},
  organization={IEEE}
}
\bibliographystyle{IEEEtran}

\end{document}